\pdfoutput=1
\documentclass{vldb}
\usepackage{graphicx}
\usepackage{balance}  
\newtheorem{problem}{Problem}
\newtheorem{definition}{Definition}
\newtheorem{example}{Example}

\newcommand{\mname}{\textsc{DRESS}}
\usepackage{amssymb,amsmath}
\usepackage{multirow}
\usepackage{mdframed}   
\newtheorem{ques}{Question}
\newmdtheoremenv[%
innerleftmargin=0.1pt,%
innerrightmargin=0.1pt,backgroundcolor=light-gray,
innertopmargin=0.1\baselineskip,%
innerbottommargin=0.1\baselineskip,%
splittopskip=0pt,skipbelow=0pt,%
skipabove=0pt,ntheorem]{myans}{Result}

\usepackage{algorithm}
\usepackage{algorithmic}
\usepackage{subfigure}
\DeclareMathOperator*{\argmin}{arg\,min}
\DeclareMathOperator*{\argmax}{arg\,max}

\newcommand{\m}{{\sf {DRESS}}}

\newcommand{\jd}{{\textit{JD.com}}}

\begin{document}


\title{Deep Reinforcement Learning \\for Personalized Search Story Recommendation}



%
%
%
%

\numberofauthors{4} 

\author{
%
%
\alignauthor
Jason (Jiasheng) Zhang\\
       \affaddr{Penn State University}\\
       \email{jpz5181@psu.edu}
\alignauthor
Junming Yin\\
       \affaddr{University of Arizona}\\
       \email{junmingy@email.arizona.edu}
\alignauthor Dongwon Lee\\
       \affaddr{Penn State University}\\
       \email{dongwon@psu.edu}
\and  
\alignauthor Linhong Zhu\\
       \affaddr{Facebook}\\
       \email{linhongz@acm.org}
}
\def\addauthorsection{\ifnum\originalaucount>6
    \section{Additional Authors}\the\addauthors
  \fi}

\maketitle
\begin{abstract}
In recent years, \emph{search story}, a combined display with other organic channels, has become a major source of user traffic on platforms such as e-commerce search platforms, news feed platforms and web and image search platforms.
The recommended search story guides a user to identify her own preference and personal intent, which subsequently influences the user's real-time and long-term search behavior. 
As search stories become increasingly important, in this work, we study the problem of personalized search story recommendation within a search engine, which aims to suggest a search story relevant to both a search keyword and an individual user's interest. 
To address the challenge of modeling both immediate and future values of recommended search stories (i.e., cross-channel effect), for which conventional supervised learning framework is not applicable, we resort to a Markov decision process and propose a deep reinforcement learning architecture trained by both imitation learning and reinforcement learning.
We empirically demonstrate the effectiveness of our proposed approach through extensive experiments on real-world data sets from {\jd}.
\end{abstract}

\section{Introduction}\label{sec:introduction}
\begin{figure}[!t]
\centering
\subfigure[Display search story within organic product item search page]{\includegraphics[width=0.45\columnwidth]{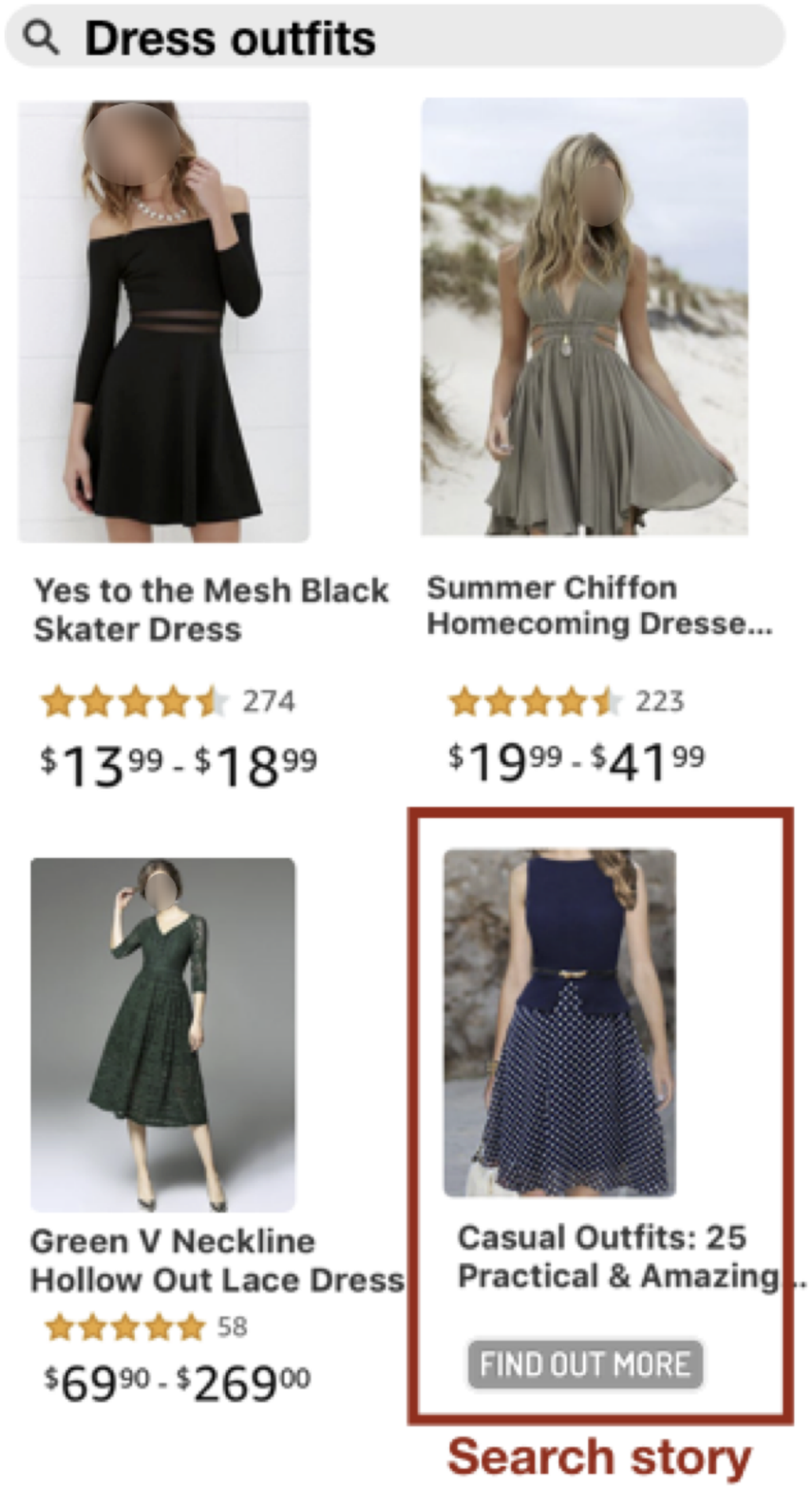}}~
\subfigure[Landing page after clicking search story, which contains both shopping guides and shopping product items]{\includegraphics[width=0.45\columnwidth]{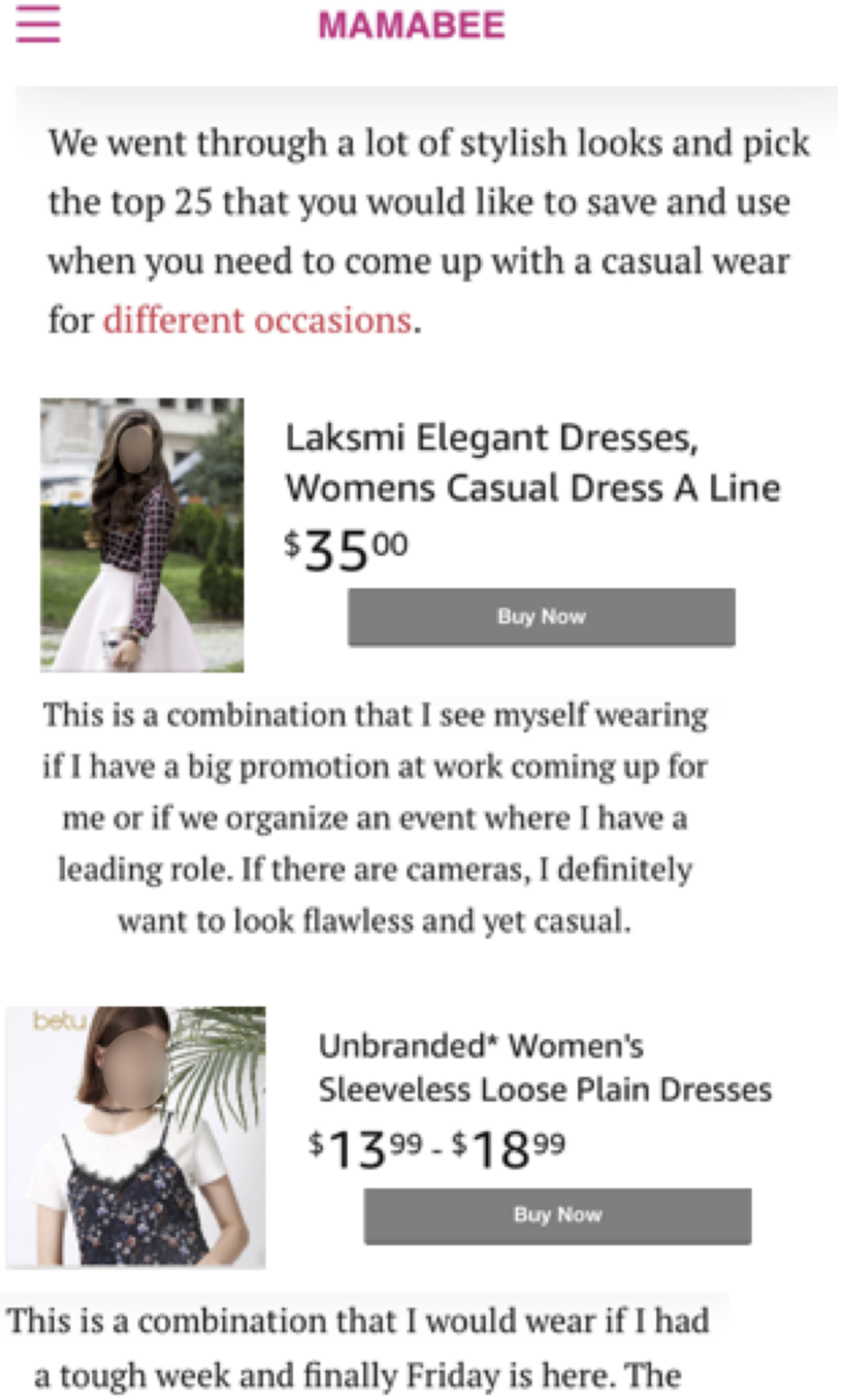}}
\caption{An illustrated (not a screenshot) example of search story recommendation.
}\label{fig:UI}
\end{figure}

Imagine that a customer visits a retail shop to purchase a dress which is to her liking. As the customer walks in, a business assistant is present to assist the customer by answering questions on fashion trend or suggesting related dresses. In online e-commerce applications, more business units are adding a component that plays a similar role as the business assistant in a shop. In this paper, we are interested in a particular component, commonly known as {\bf search story}, that has become popular among e-commerce search engines on many online platforms. For instance, in news feed platforms and web and image search platforms, each search story is a display of recommended high-quality content which is relevant to a user's personal interests. In e-commerce search platforms, on the other hand, a search story is instead a display of sponsored article that gives an overview and comparison of several product items. 
Figure~\ref{fig:UI} illustrates an example of a search story in a real e-commerce search engine, which is embedded within the organic search results. 
In this example, the search story itself displays, when clicked, a short survey that summarizes and compares a list of selected product items and related styles.

The search story recommendation can be naturally formulated as a conventional recommendation or ranking problem that aims to suggest relevant items to users based on search keywords.
For instance, one may model the problem as a click-through prediction task and recommend the search story with the highest predicted click-through rate. 
However, compared with conventional recommendation systems or search engines, recommendation of search stories focuses more on guiding users to figure out their own preferences and personal intents. 
Consider the following concrete example that illustrates a multitude of objectives of a search story recommender.

\begin{example}
As shown in Figure~\ref{fig:UI}, suppose a customer wants to purchase a ``dress outfit" for a party, but does not know what exact style she is looking for (e.g., ``sleepless loose plain dress"). 
The purpose of a search story recommender is to assist and guide the customer within each search session, as if it plays the role of an assistant in a shop. 
On the one hand, a user's search session history can be leveraged to learn the user's intent and subsequently to build a better recommendation model for future search stories. 
On the other hand, the recommended search story guides the user to figure out her preferences and personal intents, which affects not only her immediate behavior (e.g., clicking or ordering product items from the current page of the search story in Figure~\ref{fig:UI}(b)), but also her long-term behavior (e.g., clicking or ordering product items in future search session in Figure~\ref{fig:UI}(a)).
\end{example}
As this example illustrates, the ultimate goal of a search story recommendation in e-commerce search applications is to recommend the best search story that maximizes both {\em short-term} reward (e.g., purchasing a product shown in the landing page of a search story) and {\em long-term} reward (e.g., returning back to start another search session in a week). Compared with organic search results, search stories risk disrupting users' current search to achieve better long-term benefit in their following search. Therefore, search story recommendation requires a solution to consider both immediate and future benefits. 
Although we consider direct feedbacks (i.e., users' clicking or ordering product items in the landing page of search stories), indirect feedbacks (i.e., users' clicking or ordering product items in the search page) is more important.
Such a cross-channel effect~\cite{abe2004cross} is hard to model using the conventional supervised learning framework. This motivates us to propose a novel reinforcement learning framework for personalized search story recommendation.

Concretely, we formulate the personalized search story recommendation problem as a Markov decision process and propose a deep reinforcement learning architecture with 
(1) a combination of both imitation learning and reinforcement learning, as well as (2) a combination of both model-based and model-free reinforcement learning. 
Our deep reinforcement learning solution, named as {\m} (\underline{D}eep \underline{RE}inforce-\\ment learning for \underline{S}earch \underline{S}tory recommendation), consists of three components: a dynamic model parameterized by a recurrent neural network, an actor network with a proximal policy optimizer~\cite{schulman2017proximal}, and a critic network. 
The dynamic model is used to infer user behavior pattern (i.e., the environment) and is applied as the virtual environment for the controller learning. 
Such a model-based method complements model-free method for data efficiency, which is critical when only offline data is available. The actor network provides the policy (i.e., the distribution of recommended search stories) based on the state, the user' behavior history, and the current query. 
We use both imitation learning and critic network (reinforcement learning) to tune the actor network. 
The imitation learning procedure fits the actor network to the offline data, so that on one hand the stochastic logging policy of the offline data is estimated, and on the other hand, the actor network is warmed-up for further tuning. 
The critic network estimates the long-term reward (i.e., state value function or advantage) of the logging policy, which can be used to tune the actor network based on the idea of safe policy iteration.

The main contribution of this work can be summarized as follows:
\begin{description}
\item[Novel Problem.] We study an emerging search story recommendation problem and develop a solution based on deep reinforcement learning framework, addressing the challenges that originate from its cross-channel and long-term property. 
\item[Sound Methodology.] We propose an architecture combining model-free and model-based reinforcement learning, as well as imitation learning and reinforcement learning as a strategy to apply safe policy iteration to offline data.
\item[Practical Solution.] Experiments on real-life data sets from {\jd} have empirically demonstrated the effectiveness of our proposed solution.
\end{description}

The remainder of the paper is organized as follows.
We review related literature in Section~\ref{sec:related} and formulate the problem in Section~\ref{sec:problem}.
Section~\ref{sec:framework} provides a brief overview of our proposed solution.
We present the dynamic model and controller in Section~\ref{sec:dynamicmodel} and Section~\ref{sec:controller} respectively, and introduce imitation and imagination learning in Section~\ref{sec:imitation}. 
The experimental results are shown in Section~\ref{sec:expt}. 
We finally conclude our work in Section~\ref{sec:conclude}.

\section{Related Work}\label{sec:related}
In this session, we briefly review two topics that are relevant to our work, namely reinforcement learning and recommendation/ranking.

\subsection{Reinforcement Learning}
In the general reinforcement learning framework, an agent sequentially interacts with the environment and learns to achieve the best return, which is in the form of accumulated immediate rewards. 
In the partially observable Markov decision process (\texttt{POMDP}) model, at each time step $t$, when the agent has the observation of the environment $o_{t}$, an action $a_{t}$ is taken to obtain the reward $r_{t}$ from the environment. 
As the environment is partially observable, the state $s_{t}$ of the environment at time $t$ can only be inferred from the whole history up to time $t$, which can be denoted as $s_{t}=\delta (o_{1},a_{1},r_{1},...,o_{t-1},a_{t-1},r_{t-1}, o_{t})$. 
The goal of the reinforcement learning problem is to learn an optimal policy, a sequence of decisions mapping state $s$ to action $a$, to maximize the expected accumulated long term reward.

\begin{table}
\caption{Summary of representative works in deep reinforcement learning.}\label{tab:RL}
\begin{tabular}{|c|p{1.5in}|p{0.8in}|}
\hline
Element&Group&Representative\\
\hline
Value&Deep Q-learning Network (DQN)&\cite{mnih2015human,van2016deep,wang2015dueling}\\
\hline
\multirow{2}{*}{Policy}&Actor-Critic&\cite{mnih2016asynchronous,schulman2015trust,schulman2017proximal}\\
\cline{2-3}
&Deterministic Policy Gradient (DPG)&\cite{silver2014deterministic,lillicrap2015continuous,watter2015embed}\\
\hline
\multirow{2}{*}{Model}&Model-Based&\cite{sutton1990integrated,gu2016continuous}\\
\cline{2-3}
&Model-free&\cite{mnih2015human,van2016deep,wang2015dueling,mnih2016asynchronous,schulman2015trust,schulman2017proximal,silver2014deterministic,lillicrap2015continuous,watter2015embed}\\
\hline
\end{tabular}
\end{table}

Remarkably, deep reinforcement learning has achieved notable success in various tasks including but not limited to game playing~\cite{mnih2015human,silver2016mastering}, search and recommendation~\cite{theocharous2015personalized,zheng2018drn}, robotics and autonomous vehicles~\cite{levine2016end,michels2005high}, online advertising~\cite{bottou2013counterfactual, cai2017real}, several NLP tasks~\cite{li2016deep,bahdanau2016actor} and database management systems~\cite{li2018model,trummer2018skinnerdb,zhang2019an}. 
A list of representative works (though not exhaustive) is summarized in Table~\ref{tab:RL}, and we refer readers to surveys~\cite{li2017deep,arulkumaran2017brief} for more details. 
\subsection{Recommendation}
To build a high-quality recommendation system, one needs to understand and characterize the individual profile and behaviors of users, items, and their interactions. The commonly used factorization models~\cite{weimer2007maximum,koren2009matrix} learn factors for user and item by decomposing user-item interaction matrices. Neighborhood methods~\cite{sarwar2001item,koren2008factorization,guerraoui2017heterogeneous} rely on similarities between users and items that are derived from content or co-occurrence. These popular methods often ignore or under-exploit important temporal dynamics and sequential properties of the interaction between users and items.

In addition to these popular methods, deep feed-forward networks have been successfully applied in recommender systems. \cite{salakhutdinov2007restricted} used restricted Boltzmann machines for collaborative filtering and achieved remarkable results. Other feed-forward models~\cite{van2013deep,wang2015collaborative, covington2016deep} (e.g. convolutional neural networks, stacked denoising auto-encoders) have also been used to extract feature representations from items to improve recommendation.

In order to exploit the temporal dynamics and sequential information, \cite{hidasi2015session} introduced recurrent neural network (RNN) to recommendation system on the task of session based recommendation. They devised a GRU-based RNNs and demonstrated good performance with one hot encoding item input and rank based loss functions. Further improvements on session based recommendation include exploiting rich features like image \cite{hidasiparallel} and data augmentation \cite{improved16Tan}.

\subsection{Reinforcement Learning in Recommendation and Ranking}
All the above works on recommendation still focus on one round static optimization of the recommendation model. To better incorporate real-time user's feedback, several contextual bandit based ranking/recommendation approaches~\cite{radlinski2008learning,li2010contextual,zhao2013interactive} were proposed to update the selection strategy based on user-click feedback to maximize total user clicks.

However, a major assumption of bandit approaches is the ineffectiveness of action (i.e., choice of arms in bandit) on the environment state transitions,
which fails in personalized search story recommendation scenario, where the environment state or users' preference and intent here will be affected by the recommended search story. Hence we turn to reinforcement learning (RL) framework which can take into account the long-term effect of current actions.

There are some pioneering works applying RL to different tasks in recommendation and ranking, such as cross-channel recommendation \cite{abe2004cross}, personalized news recommendation~\cite{zheng2018drn}, impression allocation of advertisements~\cite{cai2018reinforcement}, and learn-to-rank for search sessions~\cite{hu2018reinforcement}. Their motivations to use RL are all based on the long-term effect of current actions in the corresponding problems. For example, in personalized news recommendation, the current recommended piece may shape the users' interests so that it can affect later recommendation results~\cite{zheng2018drn}.


\section{Problem Definition}\label{sec:problem}
\subsection{Preliminary}
For ease of presentation, we first introduce the list of notations and basic concepts used through the entire work. Specifically, we use lower case symbols $u$, $q$, $d$, $p$ to represent a single user, query, story item, and an item from another channel (e.g., the product item), respectively. 
Upper case symbols $U$, $Q$, $D$, $P$ are used to represent a set of users, queries, story items, and product items respectively. 
Let $f$ denote the search story recommendation function that maps a context $c$ to a selected story $d\in D$. 
The context $c$ can be a specific query $q$ for general search or a specific user $u$ for recommendation or a single user $u$ plus a single query $q$ for personalized search. 
With the above notations, we define the concept of \emph{search session} and \emph{personalized search episode} as follows.

\begin{definition}[\textbf{Search Session}]\label{def:session}
A search session is a series of feedback $\mathcal{I}$ (e.g., click, order, page view) by the user $u$ at time $t$ towards the returned page with a search story $d$ addressing a given query $q$. Formally, we can use a tuple $e=<t$, $u$, $q$, $d$, $\mathcal{I}>$ to denote a search session. 
\end{definition}

\begin{definition}[\textbf{Search Episode}]
A search episode $E$ is a temporal sequence of search sessions by the user $u$, which is denoted as $E=(e_1$, $\cdots$, $e_{t}$, $\cdots$, $e_{T})$. 
We add a subscript to $E$ (i.e., $E_u$) to denote a search episode of a specific user $u$.
\end{definition}



\subsection{Problem Formulation}\label{subsec:problem}
As introduced earlier, in this work we focus on  reinforcement learning for personalized search story recommendation. Specifically, we aim to find a strategy that updates the search story item recommendation function of a search engine along search episodes to achieve the best reward for each user. 

When putting the personalized search story learning into the general reinforcement learning framework, the corresponding observation $o_t$, the action $a_{t}$, the state $s_t$, the transition $\mathbb{T}$, the reward $r_t$ are defined as:
\begin{description}
\item[Observation $o_t$] is the user-dependent and query-dependent feature space $X$ for story items with each item $d$ represented as $x(u, q, d, t)$ or in short $x_t$. 
\item[Action $a_{t}$] is the selection of the search story $d\in D$.
\item[State $s_t$] is the combination of users search episode up to time $t$, i.e., the history $E^{1:\,t-1}=(e_{1}, ..., e_{\tau}, ..., e_{t-1})$, and the observation $o_t$.
\item[Transition $\mathbb{T}$] is the state transition function dependent on $a_{t}$, $s_{t+1} = \mathbb{T}(s_{t}, a_{t})$.
\item[Reward $r_t(s_t, a_t)$] can be quantified as the number of clicks, or the number of orders, or gross merchandise volume received from users when users are in state $s_t$ and search story recommender performs action $a_t$. In this work, we set reward as the binary indicator whether users click any products in the search session $e_t$.
\end{description}

Therefore, in this work, we aim to solve the following problem:
\begin{problem}\label{problem:general}
Given the entire search episode of a user $E_u$, we aim to sequentially refine the action towards each search session $e_u$ based on observed feature space $X$ and a policy $\pi(a|s)$. Specifically, at each time step $t$, the objective is to find the best policy to maximize the estimated cumulative rewards. That is:
\begin{equation}
\begin{aligned}
& \argmax_{\pi} &  & \mathbb{E}[R_1^T|\mid s_{t}, a_{t-1}, \pi]\\
& \text{subject to } & & R_1^T= \sum_{\tau=1}^{T}\gamma^{\tau}r_{\tau}(s_{\tau}, a_{\tau}),
\end{aligned}
\label{eq:return}
\end{equation}
where $R_1^T$ is the discounted cumulative rewards, $\gamma\in[0,1]$ is the discount factor, and $\mathbb{E}[x]$ denotes the expectation of $x$.
\end{problem}
\section{Deep Reinforcement Learning \\for Search Story Recommendation}\label{sec:framework}
In this section, we give an overview of our deep reinforcement learning framework for personalized search story recommendation, {\m}. Given limited offline data, we propose to combine both model-based augmentation and imitation learning with the conventional reinforcement learning. Model-based reinforcement learning requires much less training data compared to model-free reinforcement learning. The data efficiency provides additional benefits such as faster model iteration and less storage of logging data, both of which are very important for industry applications. On the other hand, imitation learning estimates the logging policy (that leads to the offline data) from the offline data, which is both the initialization of the actor network and a critical component in safe policy iteration controller learning algorithm.



\begin{algorithm}
\caption{DRL for Search Story Recommendation}\label{alg:DRL}
\begin{tabbing}
\textbf{Input}: Logging Data $\mathcal{D}_{Log}$\\
\textbf{Output}: The search story recommender\\
1: $M_{\theta}$ = \texttt{Dynamic\_Model\_Training}($\mathcal{D}_{Log}$) (Section~\ref{sec:dynamicmodel})\\
2: Initialize the critic network $\mathbb{V}_{\vartheta}$, actor network $\pi_{\Theta}$\\
\hspace{0.3cm}// imitation learning\\
3: $\pi_{\Theta}^0$ = \texttt{Controller\_Imitation}($\mathcal{D}_{Log}$, $M^T_{\theta}$) (Section~\ref{subsec:imitation})\\
\hspace{0.3cm}// one step reinforcement learning on $\mathcal{D}_{Log}$\\
4: $\pi_{\Theta}$, $\mathbb{V}_{\vartheta}$ = \texttt{Controller\_Learning}($\mathcal{D}_{Log}$, $\mathbb{V}_{\vartheta}$, $\pi_{\Theta}^0$)\\
\hspace{6cm}(Algorithm~\ref{alg:controller-RL})\\
\hspace{0.3cm}// reinforcement learning on $\mathcal{D}_{RL}$ (Section~\ref{sec:controller})\\
5: \textbf{repeat} \\
6: \hspace{0.5cm} $\mathcal{D}_{RL}$ = \texttt{Imagine}($M_{\theta}$, $\pi_{\Theta}$) (Section~\ref{subsec:imagine})\\
7: \hspace{0.5cm} $\pi_{\Theta}$, $\mathbb{V}_{\vartheta}$ = \texttt{Controller\_Learning}($\mathcal{D}_{RL}$, $\mathbb{V}_{\vartheta}$, $\pi_{\Theta}$)\\
\hspace{6cm}(Algorithm~\ref{alg:controller-RL})\\
8: \textbf{until} converges\\
9: \textbf{return} $M_{\theta}$, $\mathbb{V}_{\vartheta}$, $\pi_{\Theta}$
\end{tabbing}
\end{algorithm}

The approach is outlined in Algorithm~\ref{alg:DRL}. Randomly sampled logging search session data are collected and added to dataset $\mathcal{D}_{Log}$, which is used to train the dynamic model $M_{\theta}$ as proposed in Section~\ref{sec:dynamicmodel} (Line 1). The dynamic model serves as a virtual environment that interacts with our search story recommendation controller to learn a better recommendation policy. Search story recommendation controller is built upon the Actor-Critic framework~\cite{bahdanau2016actor}, which is parametrized as $\pi_{\Theta}$ and $\mathbb{Q}_{\vartheta}$ (Line 2). Next, instead of directly performing reinforcement learning with environment, an initial policy was learned from log data $\mathcal{D}_{Log}$ with the controller imitation learning (Line 3). We thus further improve the initial policy with a standard proximal policy gradient approach~\cite{schulman2017proximal} from the logging data $\mathcal{D}_{Log}$ (Line 4). Ideally, the controller would like to gather new on-policy data and iteratively learn a better policy in an on-policy manner. However, in this application, our ``on-policy'' data are generated by the virtual environment-- the dynamic model $M_{\theta}$. We thus repeatedly perform the above procedure to learn a better policy: 1) perform controller imagination to gather new session data and add them to a separate dataset $\mathcal{D}_{RL}$; 2) perform controller reinforcement learning to improve the recommendation policy from $\mathcal{D}_{RL}$ (Lines 5 -- 8).

\section{The Neural Network Dynamic\\ Function}\label{sec:dynamicmodel}
\subsection{Illustrative Overview}
\begin{figure*}
 \centering
 \includegraphics[width=0.8\textwidth]{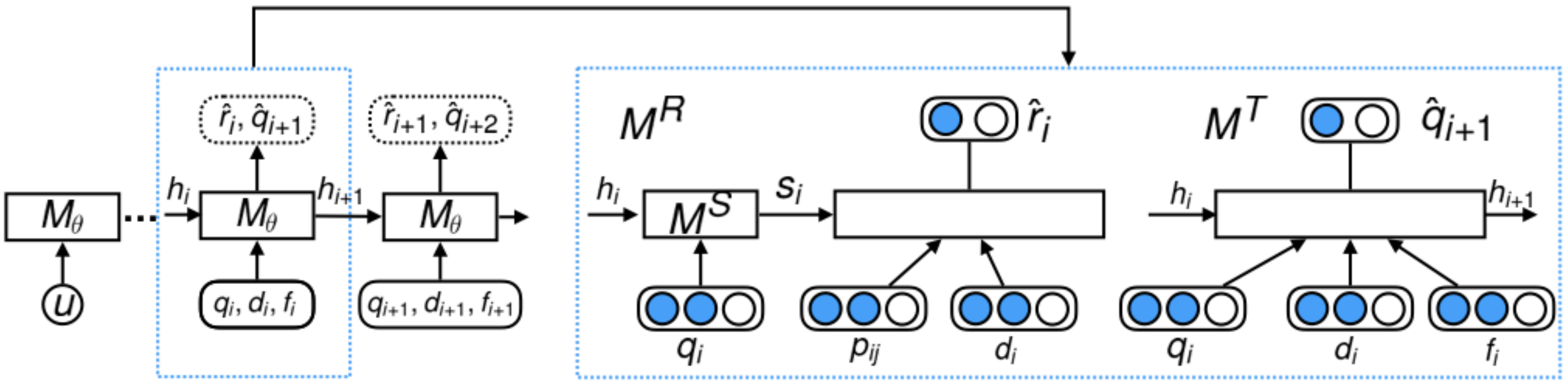}
\caption{The illustrative view of neural network dynamic function.}
 \label{fig:dynamicmodel}
\end{figure*}
As introduced earlier, we parameterize the dynamic model $M_{\theta}$ as a neural network function and thus $\theta$ represents the weights of neural networks. As illustrated in Figure~\ref{fig:dynamicmodel}, our dynamic model consists of two units: a reward model $M^R$ and a transition model $M^T$. The transition model $M^T$ updates the user hidden feature $h_i$ to $h_{i+1}$ and predicts the next query $q_{i+1}$, based on the user search session $e_i$ and user hidden feature $h_i$ as inputs. The user hidden feature $h_i$ is the hidden state of recurrently applying $M_T$ to the user search sessions until timestamp $i$ and the initial user hidden feature $h_{0}$ is determined by the user profile $u$. 

The reward model can be intuitively interpreted as a click-through prediction model (\texttt{CTR} model). 
The inputs are the user hidden feature $h_i$, the query $q_i$, the product item $p_{ij}$, and the search story $d_i$, whereas the output is the reward $\hat{r}_i$. The state submodule $M_s$ takes input $h_i$ and $q_i$ and outputs the user state $s_i$, representing the user intent. And then $s_i$ is combined with $p_{ij}$ and $d_{i}$ as inputs to the core submodule to predict the reward $\hat{r}_i$.

In the following, we introduce the detailed implementation of transition and reward model architectures including featurizations, loss function, and optimizers. Note that although we introduce them separately, these two units are implemented within the same architecture and various layers (variables) are shared. For instance, the user hidden feature $h_i$ is shared across both the transition model and reward model.

\subsection{Transition Model}
We outlined the detailed architecture of transition model on the left side of Figure~\ref{fig:dynamiccell}. 

\subsubsection{Featurization}
\begin{figure*}
 \centering
 \includegraphics[width=0.8\textwidth]{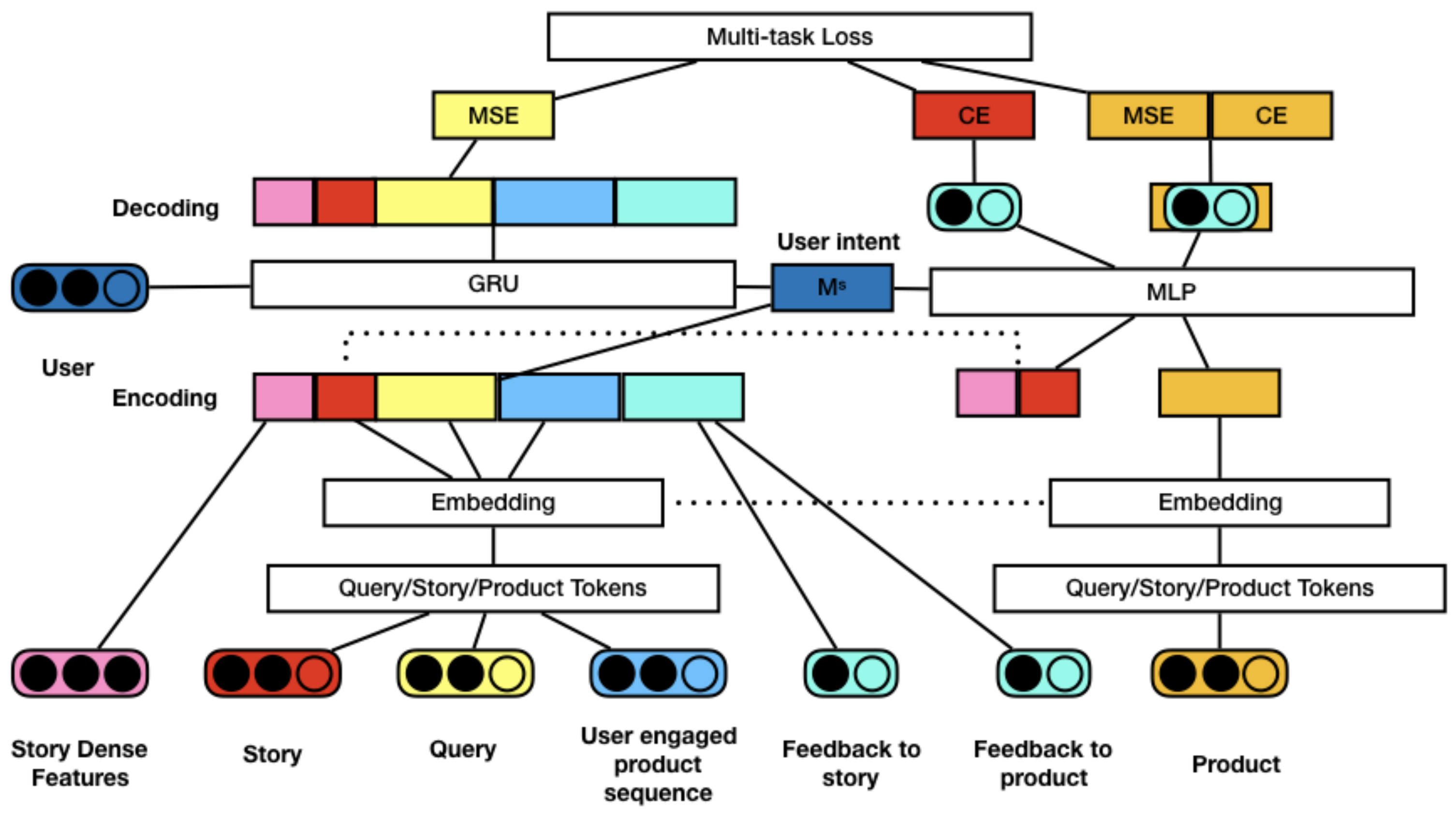}
\caption{The architecture of implemented RNN dynamic model. Colors are used to distinguish different types of objects. Components which are connected by dotted line denote shared module across transition model and reward model (best viewed in color).}
 \label{fig:dynamiccell}
\end{figure*}

The hidden feature $h_0$, is represented as a user vector constructed from both user's long-term profile and real-time profile. Regarding the user search session $e_i$, as defined in Definition~\ref{def:session}, the user search session $e_i$ consists of the query $q_i$, the story $d$, and the feedback $\mathcal{I}$. For each query, as shown in Figure~\ref{fig:dynamiccell}, we represent it as an aggregated vector of its token embeddings (yellow boxes). For each story, we first represent it as raw tokens plus dense human crafted features. The raw tokens were obtained from both the title/description of story itself as well as those of product items within the story $d$. The raw tokens were fed into the embedding layer (shared with query embedding) and transformed into an aggregated vector of token embeddings (red boxes). The aggregated embedding vector (red box) was concatenated with dense vectors (pink box) as the final representation of the story $d$. 

The feedback $\mathcal{I}$ was represented as the concatenation of two one-hot encoding session-level search story/product item engagement binary indicator vectors (green boxes) and the aggregated vector of token embeddings from user engaged product items (light blue boxes). 

\subsubsection{Layers of Model}
The transition model is empowered with a traditional encoding-decoding architecture using the gated recurrent unit (GRU). The inputs are the concatenation of feature vectors of story, query and feedbacks as well as the hidden state $h_i$. The output is the feature representation of predicted next query $q_{i + 1}$.

\subsubsection{Loss Functions}
We simply use the mean square error \texttt{MSE} between the predicted feature vector of query and the ground truth feature vector of query as the loss function for the transition model.
\begin{equation}
\mathcal{L}_{T} = \mathrm{MSE}(\hat{q_i}, q_{i+1}),
\end{equation}
where $\mathrm{MSE}(\hat{y}, y) =||\hat{y} - y||_{2}^{2}$.

\subsection{Reward Model}
The architecture of reward model is outlined on the right side of Figure~\ref{fig:dynamiccell}. 

\subsubsection{Featurization}
The featurization of search story $d$ is the same as in the transition model. For the product item $p_{ij}$, similar to the search story, we represent it as an aggregated vector of token embeddings (orange box). The user intent $s_i$ (the dark blue box), was featurized as a hidden representation, which is learned by the state submodule $M_s$. The $M_s$ takes the input of hidden history $h_i$ (shared with the transition model) and observed query $q_i$ (same featurization as the transition model, yellow box) and outputs the user state $s_i$.

\subsubsection{Layers of Model}
We simply use a multilayer perceptron (MLP) network, which takes the input of user, search story, product items and predict the feedback for search story and product items. The output layer is formulated as a classification layer for search story feedback prediction and a combination of classification and regression layer for product item feedback prediction.

\subsubsection{Loss function}
For the classification layer, we simply use the cross entropy loss (CE), while for the regression layer, we use the conditional square error (CSE). Specifically, assume that the ground truth feedback label for search story and product item and the ground truth product representation is $y_d/y_p/y_{rp}$, and the predicted feedback label for search story and product item and the product representation is $\hat{y}_d$/$\hat{y}_p$/$\hat{y}_{rp}$, the loss function is defined as:
\begin{equation}
\begin{aligned}
\mathcal{L}_{D} &= \mathrm{CE} (\hat{y}_d, y_d)\\
\mathcal{L}_{P} &= \mathrm{CE} (\hat{y}_p, y_p)\\
\mathcal{L}_{P_l} &= \mathrm{CSE} (\hat{y}_{rp}, y_{rp}|y_p)\\
\end{aligned}
\label{eq:feedbackloss}
\end{equation}
where the cross entropy loss $\mathrm{CE}$ is defined as: $\mathrm{CE} (\hat{y}, y) = - y \log \hat{y} - (1 - y) \log (1 -\hat{y})$ and the conditional square error $\mathrm{CSE}$ is defined as: $\mathrm{CSE}(\hat{y}_{rp}, y_{rp}|y_p)=y_p||\hat{y}_{rp}-y_{rp}||_{2}^{2}$.

\subsection{Dynamic Model Training}\label{subsec:model-train}
Given the logging data $\mathcal{D}_{log}$, we thus train the dynamic model by optimizing the following loss function:
\begin{equation}
\mathcal{L}_M = w_T \mathcal{L}_{T}  + w_D \mathcal{L}_{D} + w_P \mathcal{L}_{P} + w_{P_l} \mathcal{L}_{P_l}
\label{eq:modelloss}
\end{equation}
where $w$ is the coefficient that is proportional to the contribution of each loss function. 
For ease of presentation, we use $(M^R_{\theta}, M^T_{\theta})$ = Dynamic\_Model\_Training($\mathcal{D}_{Log}$) to denote the procedure of training the dynamic model with the architecture shown in Figure~\ref{fig:dynamiccell}.

\section{Controller Reinforcement\\ Learning}\label{sec:controller}
\begin{figure*}[tb]
 \centering
 \includegraphics[width=0.8\textwidth]{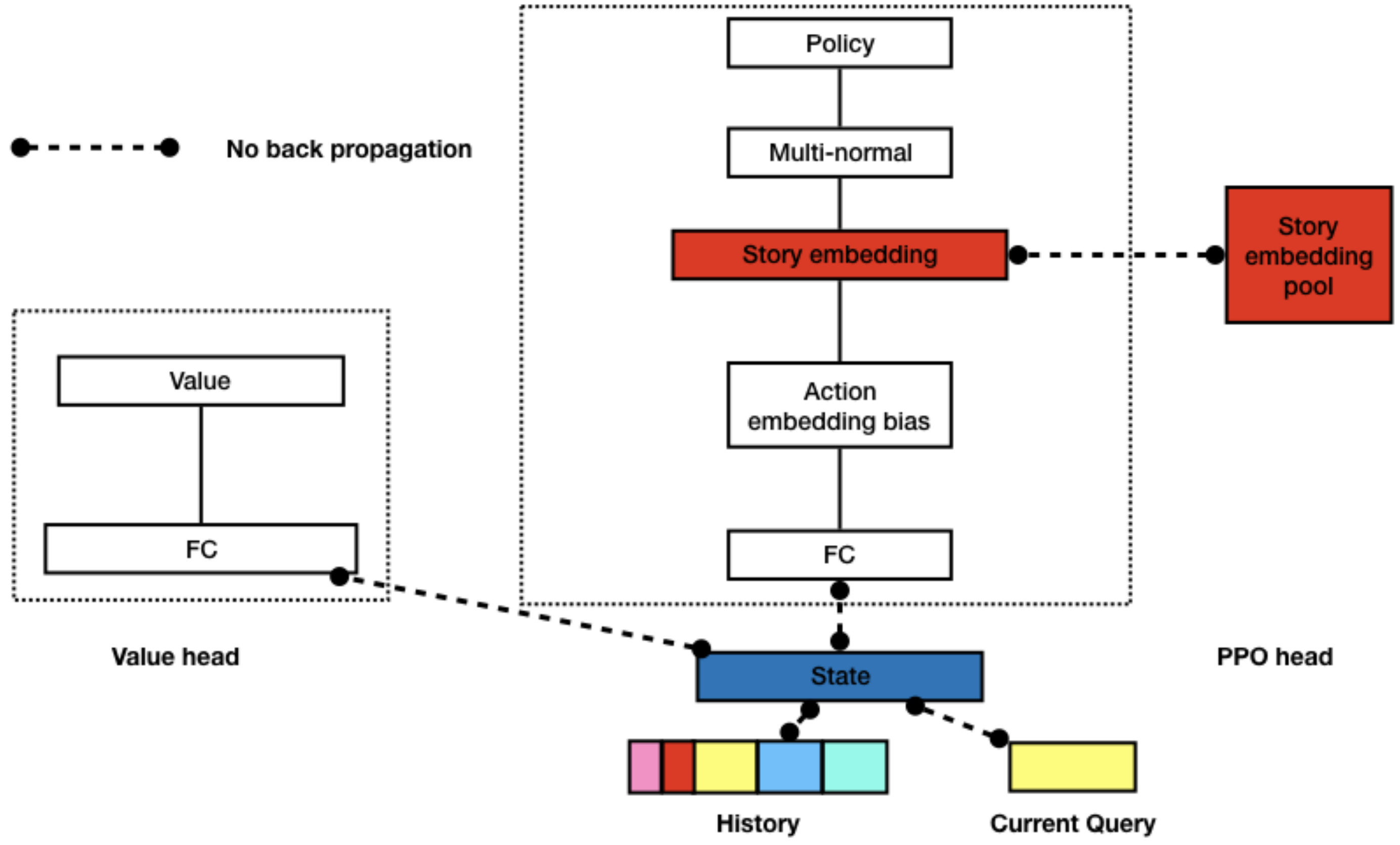}
\caption{Network structure of reinforcement learning controller (best viewed in color).}
 \label{fig:RLController}
\end{figure*}
Our reinforcement learning controller is designed under the traditional actor-critic architecture~\cite{bahdanau2016actor}. Specifically, the controller is a multi-head neural network, which is used as the function approximator for choosing the best story from the story embedding pool. Figure~\ref{fig:RLController} illustrates our network structure of reinforcement learning controller, which consists of the state-value head (i.e., critic network) and policy head (i.e., actor network) with the shared input of state representation, the user hidden feature $h_i$ from the transition model $M^{T}$. The details are presented as follows.
\subsection{Critic Network}\label{subsec:critic}
As shown in Figure~\ref{fig:RLController}, the value network is joinly learned with the policy network, where the input is the user hidden feature $h_t$ from the transition model $M^{T}$, representing the state $s_t$, and the output is the $\mathbb{V}_{\pi}$ value $\mathbb{V}_{\pi}(s_t)$ of state $s_t$ under policy $\pi$. Without ambiguity, we use $\mathbb{V}$, omitting the policy subscription. Our value network uses a neural network to learn the value function $\mathbb{V}$ with parameter $\vartheta$. Specifically, the $\vartheta$ is updated by the gradient descent optimizer with the following loss function:
\begin{equation}\label{equ:valueloss}
\begin{aligned}
\mathcal{L}_{\vartheta}^{\mathbb{V}}(\pi_{\Theta}) = \mathrm{MSE}(\mathbb{V}_{\vartheta}(s_t), \mathbb{V}^{\texttt{target}}(s_t))\\
\mathbb{V}^{\texttt{target}}(s_t) = r(s_t, a_t) + \gamma \mathbb{V} (s_{t+1})
\end{aligned}
\end{equation}

The updated formula of the parameter $\vartheta$ with regard to Equation~\ref{equ:valueloss} is the stochastic version of the Bellman equation.
\subsection{Actor Network}\label{subsec:actor}
Our policy optimization is designed based on the state-of-the-art Proximal Policy Optimization (PPO) controller~\cite{schulman2017proximal}. Our policy $\pi$, is again parametrized as a neural network function with parameter $\Theta$ (in order to distinguish with the dynamic model parameter $\theta$). The architecture of our policy neural network, is shown on the right side of Figure~\ref{fig:RLController}.

\begin{algorithm}
\caption{\texttt{Controller\_Learning}($\mathcal{D}$, $\mathbb{V}_{\vartheta}$, $\pi_{\Theta}$)}\label{alg:controller-RL}
\begin{tabbing}
\textbf{Input}: Data $\mathcal{D}$, current actor network $\pi_{\Theta}$\\
\hspace{1.05cm}and the critic network $\mathbb{V}_{\vartheta}$\\
\textbf{Output}: The updated actor network $\pi_{\Theta}$ and \\
\hspace{1.05cm}critic network $\mathbb{V}_{\vartheta}$\\
1: \textbf{Repeat} sampling a mini-batch $bs$ of search sessions \\
\hspace{1.1cm}from $\mathcal{D}$\\
2: \hspace{1.0cm} update the critic network $\mathbb{V}_{\vartheta}$ minimizing eq.~\ref{equ:valueloss}\\
3: \hspace{1.0cm} update the actor network $\pi_{\Theta}$ minimizing eq.~\ref{equ:actorloss}\\
4: \textbf{return} $\mathbb{V}_{\vartheta}$, $\pi_{\Theta}$
\end{tabbing}
\end{algorithm}
In the controller reinforcement learning procedure, it learns the policy $\pi$ by maximizing the accumulated state value of a policy averaging over the state distribution of the search session history:
\begin{equation}\label{equ:actorloss}
\begin{aligned}
& \pi = & & \argmax_{\Theta}  \mathcal{L}_{R}(\pi_{\Theta})\\
& \mbox{subject to } & & \mathcal{L}_{R} (\pi_{\Theta})=\sum_{u}\sum_{e\in E_u}[\mathcal{L}_{e}^{\texttt{clip}}(\pi_{\Theta}) + w_{\mathbb{H}_l}\mathbb{H}(\pi_{\Theta}(\cdot|s))],
 \end{aligned}
\end{equation}
where $L_{e}^{\texttt{clip}}$ $=\min\{\frac{\pi_{\Theta}(\cdot|s)}{\pi_{\texttt{old}}(\cdot|s)}\widehat{A}, \texttt{clip}(\frac{\pi_{\Theta}(\cdot|s)}{\pi_{\texttt{old}}(\cdot|s)}, 1-\epsilon, 1+\epsilon)\widehat{A}\}$; $\widehat{A}$ is the estimated advantage function defined as $\widehat{A}_t=r_t + \gamma\mathbb{V}(s_{t+1}) -\mathbb{V}(s_t)$. The advantage function estimator here is the same with setting $\lambda=0$ in the GAE estimate for advantage used in the original PPO paper~\cite{schulman2017proximal} as the experiment suggests no better performance with a non-zero $\lambda$ value. $\mathbb{H}$ is the entropy of the policy $\pi_{\Theta}$ given state $s$, and $w_{\mathbb{H}_l}$ is the weight.

\section{Imitation and Imagination}\label{sec:imitation}
\subsection{Imitation Learning}\label{subsec:imitation}
In our search recommendation task, and most other real-world decision-making problems (e.g., finance and health-care), we have access to the logging data of the system being operated by its previous
controller, but we do not have access to an accurate simulator
of the system. The goal of the imitation learning is thus to learn to imitate the previous controller with a fixed policy $\pi_0$. 

Specifically, we learn the policy $\pi_0$ from $\mathcal{D}_{Log}$ by optimizing the likelihood of any action chosen. Formally, imitation learning can be formulated as the below optimization task:
\begin{equation}
\pi_0  = \argmin_{\Theta}  \mathcal{L}_{I}(\pi_{\Theta}),\\
\end{equation}
where $\mathcal{L}_{I}(\pi_{\Theta})$ is the likelihood function of observing actions in $\mathcal{D}_{Log}$ given the policy $\pi_{\Theta}$, together with an entropy penalty,
\begin{equation}
\mathcal{L}_{I} = 
-\sum_{u}\sum_{e\in E_u}\log(\pi_\Theta(a|s)) - w_{\mathbb{H}_I}\mathbb{H}(\pi_\Theta(\cdot|s)),
\label{eq:controllerimitation}
\end{equation}
where $w_{\mathbb{H}_I}$ is the weight for the entropy regularizer $\mathbb{H}$ .
\subsection{Controller Imagination}
\label{subsec:imagine}
It is not data efficient to only apply model-free reinforcement learning method on the logging data, especially the previous controller reinforcement learning (Section~\ref{sec:controller} )is simply one iteration of the PPO algorithm~\cite{schulman2017proximal}. The goal of controller imagination is thus to use the trained dynamic model to further improve the actor network.

Specifically, we use randomly selected sessions in $\mathcal{D}_{Log}$ as starting sessions, from each of which, the dynamic model ($M_{\theta}^{R}$, $M_{\theta}^{T}$) and the current actor network are applied to rollout $T_{img}$ fictional search sessions, stored in $\mathcal{D}_{RL}$. The imagined data $\mathcal{D}_{RL}$ is then used in the controller reinforcement learning (Section~\ref{sec:controller}) to further tune the actor network.

Generally, it is similar to the original PPO controller learning~\cite{schulman2017proximal}, except that the real environment is replaced by the dynamic model here.

\section{Experimental Validation}\label{sec:expt}
In this section, we conduct extensive experiments with a dataset from a real e-commerce company and evaluate the effectiveness of {\m}.

\subsection{Experimental Setup}
\subsubsection{Dataset} 
\begin{figure*}[tb]
\centering
\subfigure[]{\label{figure:histuser}\includegraphics[width=0.45\textwidth]{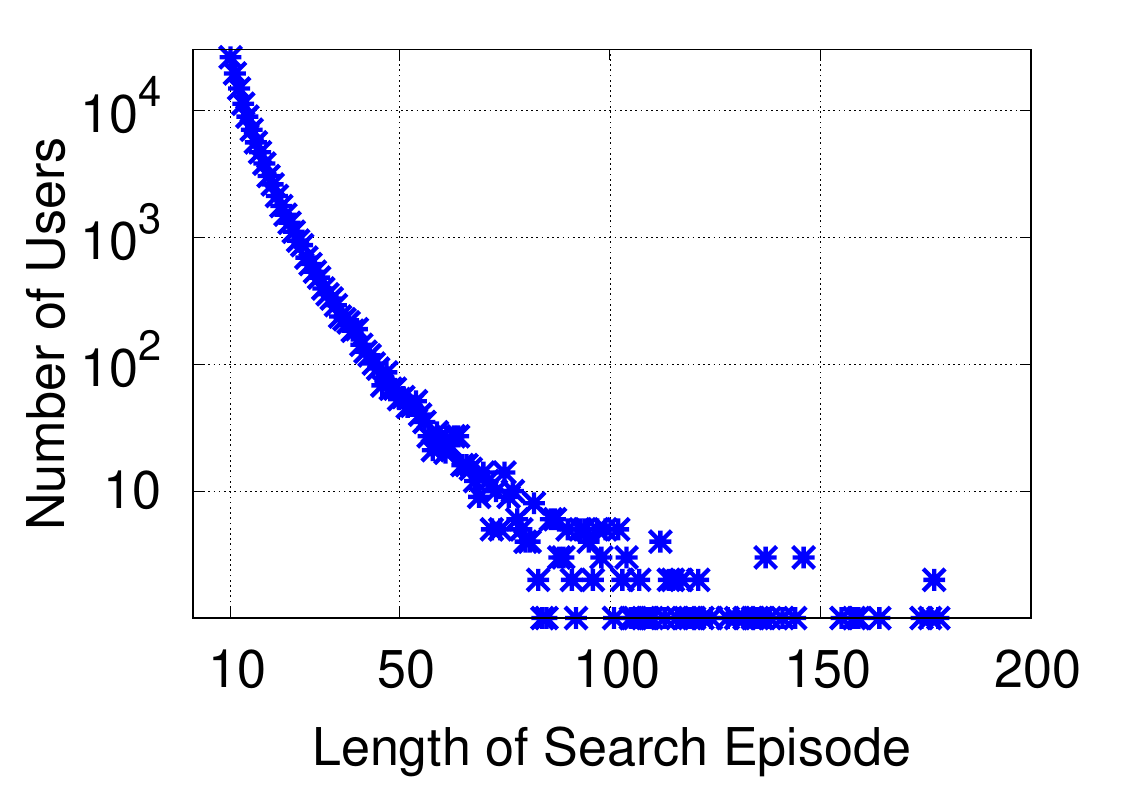}}
\subfigure[]{\label{figure:histstory}\includegraphics[width=0.45\textwidth]{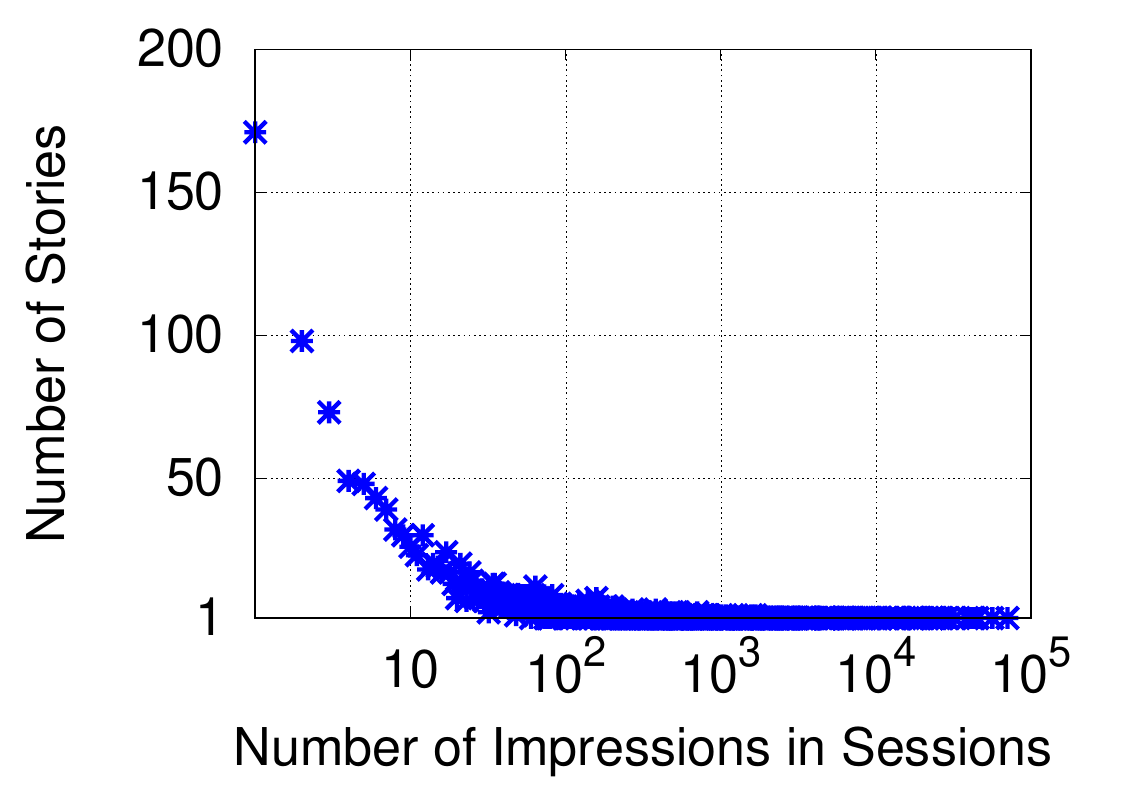}}
\vspace{-0.1in}
\caption{
Histograms of (a) episode length
and 
(b) story impression frequency. Both follow a power-law distribution.}
\end{figure*}

We evaluate our methods on a dataset collected between Apr 2018 and Jul 2018 from {\jd}~\cite{zou2019reinforcement}. 
We sampled all search sessions that are related to a category ``women dress" and filtered out search episodes with only a few sessions or a huge number of sessions. Our dataset are carefully pre-processed and anonymized. The distributions of the episode length and the number of search sessions in which each search story appears are visualized in Fig.~\ref{figure:histuser} and~\ref{figure:histstory}. As shown in Fig.~\ref{figure:histuser}, we only keep the search episodes whose length is within the range [11, 200].
Other statistics of our dataset is summarized in Table~\ref{table:data}. We randomly divide the dataset into 5 folds by users. Hence each fold of the dataset contains equal number of complete search episodes. We did 5-fold cross-validation experiments with one random fold as testing data for each experiment.

The processed feature dimensions are summarized as follows.
Each query is represented as an aggregation of 200 dimensional word embedding vectors of segmented query words. Each product is represented as an aggregation of 200 dimensional word embedding vectors of words from product titles. For each story, it is featured as a concatenation of 200 dimensional word embedding vectors of words from story titles, 200 dimensional word embedding vectors of title words from the products embedded within a story, and 13 human crafted features of a story.
\begin{table}
\caption{Statistics of dataset.}
\centering
\label{table:data}
\begin{tabular}{cccc}
\hline
\# users & \# stories & \# products & \# sessions\\
\hline
122\,,886	& 2\,,185	& 304\,,780 & 1\,,842\,,879\\
\hline
\end{tabular}
\end{table}


\subsubsection{Comparable Methods}
We compare the proposed method {\m} 
as described in Algorithm~\ref{alg:DRL} with the following baseline methods:
\begin{enumerate}
\item ORIGIN: This is the state-of-the-art implementation of a search story recommendation, 
that results in the offline data, 
currently being used by the company.

\item DNNC (Deep Neural Network Classifier): Without considering the cross-channel effect, this method is trained to recommend a search story that is likely to be clicked, given the story feedback data. To be a fair comparison, DNNC uses the  architecture with the actor network and is initialized with the imitation policy $\pi_{\Theta}^{0}$, same as \mname.

\item \mname-m: This is the myopic version of {\mname} that only considers immediate short-term reward, which is implemented by setting $\gamma=0$.

\item \mname-s: This is the simplified version of {\mname} 
with the controller imagination module (Section.\ref{subsec:imagine}) removed.
\end{enumerate}

\subsubsection{Evaluation Metric}
\label{sec:evaluationmetric}
The goal of a search story recommendation is to facilitate users during the search of products. 
Therefore, 
we use search session based user feedback on products as the main performance measure. In particular, we use the percentage of search sessions in which users have clicked a product, $CTR$ (Click Through Rate):
\begin{equation}
    \mathrm{CTR}=\frac{\sum_{e\in E}clk_e}{|E|}
\end{equation}
where $clk_e$ is a binary indicator whether a user clicked a product in a  search session $e$, which is the same as the reward $r_e$ in the RL framework. Similarly, we also use $CVR$ (Conversion Rate):
\begin{equation}
    \mathrm{CVR}=\frac{\sum_{e\in E}ord_e}{|E|}
\end{equation}
where $ord_e$ is a binary indicator whether a user ordered a product in a  search session $e$.

It is risky to evaluate the learned policy on a real-life system. Therefore, we use a statistical estimate method, Truncated Weighted Importance Sampling (TWIS), to estimate the performance from the offline test data as follows:
\begin{equation}
    \widehat{R(\pi)} = \frac{\sum_{e}\sum_{t=T_e-H}^{T_e}r_{t}\prod_{i=T_e-H}^{T_e}\frac{\pi_i}{b_i}}{\sum_{e}\sum_{t=T_e-H}^{T_e}\prod_{i=T_e-H}^{T_e}\frac{\pi_i}{b_i}}
\label{eq:evaluate}
\end{equation}
where $e$ is an episode, $H$ is the horizon for the latest sessions to use per episode, $r_{t}$ can be $clk_{t}$ ($ord_{t}$) so that $\widehat{R(\pi)}$ is the estimate of $CTR$ ($CVR$, respectively), $\pi_i=\pi(a_i|s_i)$ is the probability of the observed action given by the evaluated policy $\pi$ and $b_i=b(a_i|s_i)$ by the logging policy. With only offline data, the imitation policy $\pi_{\Theta}^{0}$ is used as the logging policy. It is justified by the following factors:
\begin{itemize}
\item All compared methods share the same imitation policy $\pi_{\Theta}^{0}$ in initialization.
\item The imitation policy $\pi_{\Theta}^{0}$ is trained to fit the offline data generated by the logging policy.
\item Following the idea of importance sampling, $\widehat{R(\pi)}$ is the estimate of $CTR$ ($CVR$) of the weighted policy $\pi^{w}=b^{*}\cdot\frac{\pi}{b}$, where $b^{*}$ is the true logging policy, which is valid when $\pi$ and $b$ are close enough. When the true logging policy $b^{*}$ is different from $b$, in an application, the policy ratio $\frac{\pi}{b}$ can still play the role as reinforcement learning complements $b^{*}$ and gets evaluated by $\widehat{R(\pi)}$.
\end{itemize}

This evaluation metric is invariant toward the arbitrary constant scale  of $b_i$ and of $\pi_i$. The truncated setting encourages the equal importance of users with the episodes of different lengths.

One potential downside of importance sampling based evaluation methods is the large variance \cite{mandel2014offline} when the target policy $\pi$ and the logging policy $b$ are very different. However, in our case with safe policy iteration and/or limited model-based improvement, both $\pi$ and $b$ are close. The exception is DNNC, which is a supervised learning method, so that there is no guarantee that the trained classifier policy will be similar enough to the imitation policy at initialization. In order to obtain valid result regarding TWIS, we add KL-divergence regularization to the negative likelihood loss of DNNC. 

\subsubsection{Hyperparameters}
Most hyperparameters are tuned using the validation set for each experiment. For reproducibility of our experimental results, values of important hyperparameters are summarized in Table.~\ref{table:hyperparameters}.

\begin{table}
\caption{Hyperparameters.}
\label{table:hyperparameters}
\resizebox{\linewidth}{!}{
\begin{tabular}{cc}
\hline
hyperparameters		& setting\\
\hline
discount factor $\gamma$ (Eq.\ref{eq:return})	& 0.7\\
transition loss weight $w_T$ (Eq.\ref{eq:modelloss})	& 1.0\\
story loss weight $w_D$ (Eq.\ref{eq:modelloss})		& 1.0\\
product CE loss weight $w_P$ (Eq.\ref{eq:modelloss}) 	& 1.0\\
product CSE loss weight $w_{P_l}$ (Eq.\ref{eq:modelloss})	& 1.0\\
Entropy weight for controller learning $w_{\mathbb{H}_l}$ (Eq.\ref{equ:actorloss})  & 0.01\\
Entropy weight for controller imitation $w_{\mathbb{H}_I}$ (Eq.\ref{eq:controllerimitation}) 	& 0.0001\\
clipping factor $\epsilon$ (Eq.\ref{equ:actorloss})   & 0.2\\
evaluation horizon $H$ (Eq.\ref{eq:evaluate})  & 15\\
\hline
\end{tabular}
}
\end{table}

\subsection{Empirical Results}
In this section, we conduct different groups of experiments to empirically validate the proposed approaches. Specifically, we aim to answer the following questions: (1) Is it necessary to formulate the search story recommendation problem as a reinforcement learning problem?; (2) Does the model-based reinforcement learning lead to a better performance of search story recommendation?; and (3) What is the advantage of combining both imitation learning and reinforcement learning?

\begin{figure}[!t]
\centering
\includegraphics[width=\linewidth]{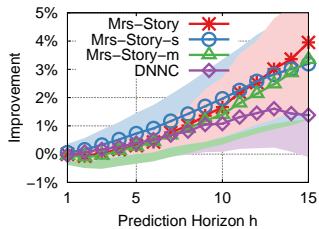}
\caption{$CTR$ Improvement versus different choice of evaluation horizon $H$. The shadow area shows the standard deviation of the multiple experiments. The points with non-integer $H$ are interpolated for better visualization.}
\label{figure:horizon}
\end{figure}

\vspace{2mm}
\hspace{-0.2in}
\fbox{\parbox[c]{\linewidth}{
\begin{ques}
\textbf{Justification of Reinforcement Learning}: Is it necessary to use the reinforcement learning framework to solve the personalized search story recommendation problem?
\end{ques}
}}
\vspace{2mm}
\begin{table}
\caption{CTR: shown as the improvement percentage over ORIGIN. 
$*$ indicates statistical significance (p-value $<0.05$)
}
\centering

\label{table:ctr}
\begin{tabular}{cc}
\hline
method	& improvement\%\\
\hline
DNNC	& 1.723\\
\mname-m	& 2.115* \\
\mname-s    & 2.546* \\
\mname		& \textbf{2.843}* \\
\hline
\end{tabular}
\end{table}

\begin{table}
\caption{CVR: shown as the improvement percentage over ORIGIN.
}
\centering
\label{table:cvr}
\begin{tabular}{cc}
\hline
method	& improvement\%\\
\hline
DNNC	& -4.374\\
\mname-m	& 15.463 \\
\mname-s    & 3.507 \\
\mname		& \textbf{19.775} \\
\hline
\end{tabular}
\end{table}
As shown in Table~\ref{table:ctr}, DNNC performs the worst compared to all other methods. DNNC is also the  method that cannot obtain significant improvement over the ORIGIN. Furthermore, from Table~\ref{table:cvr}, DNNC even performs worse than the ORIGIN in CVR. Compared with other methods, DNNC only considers the direct feedback for search stories, which ignores the feedback on organic search from another channel. Hence, this result highlights the necessity to consider the \textbf{cross-channel} effect for effective search story recommendation, as {\mname} does.
In addition, from Table~\ref{table:ctr} and Table~\ref{table:cvr}, \mname-m performs much worse than \mname. It is as expected because \mname-m only considers short-term rewards while ignoring the influence of current actions (i.e., engagement towards recommended search story) on users' long-term behavior. The result strongly suggests that one should take the \textbf{long-term} effect into consideration for effective search story recommendation.

As discussed in Section~\ref{sec:introduction}, the reinforcement learning framework is a perfect fit for supporting both cross-channel effect and long-term effect. The empirical results further suggests the strong justification to formulate the search story recommendation as a reinforcement learning problem.

\vspace{2mm}
\hspace{-0.2in}
\fbox{\parbox[c]{\linewidth}{
\begin{ques}
\textbf{Model-based versus Model-free}: Does the model-based controller imagination help improve the performance?
\end{ques}
}}

\vspace{2mm}
From Table.~\ref{table:ctr} the superiority of {\mname} 
over \mname-s shows the contribution of the model-based controller imagination sub-module, which is even more obvious in $CVR$ as shown in Table.~\ref{table:cvr}. In addition, we also show the improvement rate with different values of evaluation horizon up to $H=15$ in Figure.~\ref{figure:horizon}. It shows that the model-based controller imagination module decreases the short-term performance (i.e., smaller $H$), but increases the long-term performance (i.e., larger $H$). Hence, considering long-term performance, infinite in the real-world situation, the combination with the model-based sub-module is expected to produce better results.

\vspace{2mm}
\hspace{-0.2in}
\fbox{\parbox[c]{\linewidth}{
\begin{ques}
\textbf{Imitation + Safe policy improvement}: what is the advantage of using the safe policy iteration reinforcement learning algorithm? 
\end{ques}}}

\vspace{2mm}
The imitation policy is the estimation of current online policy that generates offline data. We expect the resulted policy to be close to the latter to ensure the stability of an online system. It is similar, if the policy ratio $\frac{\pi}{b}$ is applied to weight the current online policy $b^{*}$ to $\pi^{w}=b^{*}\frac{\pi}{b}$, as argued in Sec.\ref{sec:evaluationmetric}. We calculate three measures of distribution difference. 
\begin{enumerate}
\item Log probability ratio: $ration_i=\log(\frac{\pi(a_i|s_i)}{b(a_i|s_i)})$ for a session $i$;
\item Total variation divergence: $D_{\text{TV}}(b||\pi)_i=\frac{1}{2}\sum_{a'}|\pi(a'|s_i) - b(a'|s_i)|$~\cite{schulman2015trust}
\item KL-divergence: $D_{\text{KL}}(b||\pi)_i=\sum_{a'}b(a'|s_i)\log(\frac{b(a'|s_i)}{\pi(a'|s_i)})$.
\end{enumerate}
We calculate the averages of each difference measure over sessions in test data. We use the uniform distribution $unif$ for comparison. 
Results are shown in Table.\ref{table:policydifference}. Compared with uniform policy $unif$, both {\m} and \mname-s are close to the imitation policy. As expected, the policy obtained by {\m} deviates more from the imitation policy compared with \mname-s because of the additional controller imagination sub-module (Section.~\ref{subsec:imagine}). Hence in an application, a trade-off should be made between  performance gain and  stability by controlling how much controller imagination should be included (limiting the number of iterations of step 6,7 in Algorithm~\ref{alg:DRL}).

\begin{table}
\caption{Policy difference compared with the imitation policy.}
\centering
\label{table:policydifference}
\begin{tabular}{c|ccc}
\hline
policy 	& ratio  & $D_{\text{TV}}$ & $D_{\text{KL}}$\\
\hline
$unif$	& 6.042 	& 0.967 	& 5.959 \\
\mname	& 0.0827	& 0.0410	& 0.00679 \\
\mname-s  & 0.0488   & 0.0235   & 0.00232 \\
\hline
\end{tabular}
\end{table}
\section{Conclusion}\label{sec:conclude}
Deep reinforcement learning has been successfully used as a powerful method to capture a wide variety of non-trivial user behavior on online platforms (e.g., news feed recommendation, e-commerce search). In this work, following these successes, we applied the reinforcement learning framework to the challenging problem of cross-channel search story recommendation 
by resorting it into a Markov decision process. We further proposed a unified deep learning architecture employing both imitation learning and reinforcement learning. 
Comprehensive empirical validation 
indicates that our proposal, {\m}, is effective in improving a conversion rate on real-world data sets from {\jd}.

\balance

\bibliographystyle{abbrv}
\bibliography{ranking,RL}  




\end{document}